\documentclass[letterpaper,10pt,conference]{ieeeconf}

\IEEEoverridecommandlockouts
\usepackage{graphicx}
\usepackage{amsmath}
\usepackage{amssymb}
\usepackage{booktabs}
\usepackage{makecell}
\usepackage{multirow}
\usepackage{url}
\usepackage[hidelinks]{hyperref}
\usepackage{xspace}
\newcommand{\method}{\texttt{\textbf{UniIntervene++}}\xspace}

\title{\LARGE \bf
UniIntervene++: An Adaptive Intervention Agent for Efficient Real-World Reinforcement Learning
}

\author{
Yudong~Lin$^{1*}$,
Haoyuan~Deng$^{1*}$,
Zhuoxuan~Yuan$^{1}$,
Zaijia~Yang$^{1}$,
Yuanjiang~Xue$^{1}$, 
and~Ziwei~Wang$^{1\dagger}$%
\thanks{$^{*}$Equal contribution}
\thanks{$^{1}$ School of Electrical and Electronic Engineering, Nanyang Technological University, Singapore}
\thanks{$^{\dagger}$Corresponding author: \textit{ziwei.wang@ntu.edu.sg}
}
}

\begin{document}

\maketitle
\thispagestyle{empty}
\pagestyle{empty}

\begin{abstract}
Online reinforcement learning (RL) enables robot policies to improve through physical interaction, but the assistance they require changes as their competence evolves. Existing intervention strategies based on offline estimates or fixed decision rules can therefore become mismatched to the current policy. To address this, we propose \method, an adaptive intervention agent that learns to allocate control between autonomous execution and heterogeneous assisted behaviors during online RL. Specifically, \method first formulates the evolving RL policy, trajectory correction, and a task-structured CodePolicy as Options in a unified semi-Markov decision process and learns their relative values online. Building on this, competence-adaptive intervention periodically probes the RL policy through unassisted execution, keeping control allocation responsive to its evolving capability. Finally, coupled experience learning allows assisted behaviors to improve the RL policy, whose evolving outcomes in turn reshape future intervention decisions. In this way, \method jointly determines when to intervene, how to intervene, and when to return control as the RL policy improves. 
Across five real-world manipulation tasks, \method{} achieves an average success rate of 89.67\%, outperforming all baselines by at least 6 percentage points, while reducing human intervention to 0.77\%, a relative reduction of at least 94.6\% from the best baseline. Code is available in our
\href{https://github.com/dannyyudong/An-Adaptive-Intervention-Agent-for-Efficient-Real-World-Reinforcement-Learning}
{GitHub repository}.
% Across five real-world manipulation tasks, \method achieves 89.65\% average success while reducing the human intervention rate to 0.77\%.

% Online reinforcement learning (RL) enables robot policies to improve through physical interaction, but the assistance they require changes as their competence evolves. Intervention strategies based on offline estimates or fixed decision rules can therefore become mismatched to the current policy. We introduce \method, an adaptive intervention agent that learns a unified control-allocation policy over autonomous execution and heterogeneous assisted behaviors during online RL. We formulate this allocation as an availability-masked semi-Markov decision process, where at each Option boundary the agent selects the evolving RL policy, demonstration-guided trajectory correction, or a task-structured CodePolicy. Their relative values are continually updated from online outcomes while accounting for assistance and execution-duration costs. Adaptive RL probing maintains direct estimates of the RL policy's competence, allowing control allocation to shift naturally from assistance-heavy to autonomy-heavy execution as learning progresses. Thus, when to intervene, which intervention to use, and when to return control arise from a single learned decision process. Across five real-world manipulation tasks, \method{} achieves an average success rate of 89.65\%, outperforming all baselines by at least 5.98 percentage points, while reducing human intervention to 0.77\%, a relative reduction of at least 94.6\% from the best baseline.

\end{abstract}

%%%%%%%%%%%%%%%%%%%%%%%%%%%%%%%%%%%%%%%%%%%%%%%%%%%%%%%%%%%%%%%%%%%%%%%%%%%%%%%%
\section{Introduction}
% Online reinforcement learning (RL) offers a route to robot manipulation
% policies that improve from their own physical experience and adapt to contact
% dynamics, perception errors, and task variation. The same interaction that
% makes this adaptation possible is also expensive: rewards are often sparse,
% exploration can be unsafe or unproductive, and failed rollouts consume robot
% time. Prior data substantially improves sample efficiency in off-policy RL
% \cite{ball2023rlpd}, while human-in-the-loop systems such as HIL-SERL use timely
% operator corrections to expose the learner to useful recovery states
% \cite{luo2025hilserl}. These systems make real-world RL practical, but require a
% skilled operator to monitor training and repeatedly allocate control between
% the learning policy and corrective behavior.
Online reinforcement learning (RL) enables robot manipulation policies to
improve through physical interaction and adapt to contact dynamics, perception
errors, and task variation~\cite{lei2026performant,deng2026facet, deng2025survey}. However, real-world online RL remains costly:
rewards are often sparse, exploration can be unsafe or unproductive, and failed
rollouts consume robot time~\cite{padalkar2024guiding,schwarke2023curiosity,li2026failure}. Human-in-the-loop RL (HiL-RL), such as HIL-SERL,
mitigates these issues through timely human corrections during online learning~\cite{luo2025hilserl}. Yet current HiL-RL systems require a skilled operator to
continuously monitor training and repeatedly allocate control between autonomous
execution and assisted behavior, limiting real-world scalability~\cite{he2025uncertainty,sharma2023self, guo2026efficient}.

% Recent work has begun to automate this role. AutoSERL reuses a single
% demonstration to guide exploration and recover from predefined failure regions
% \cite{liu2026autoserl}; UniIntervene detects sustained value degradation and
% invokes a goal-conditioned recovery policy \cite{deng2026uniintervene}. In
% particular, UniIntervene formulates intervention learning as an offline
% value-based problem. Such methods provide useful automated assistance, but the
% relative utility of autonomy and assistance is not stationary during online
% training. Early in learning, structured intervention can avoid repetitive
% failures and produce useful experience. Later, excessive assistance can waste
% intervention effort and suppress the autonomous rollouts needed to reveal and
% improve the RL policy's current competence. Offline value estimates, fixed
% takeover rules, and fixed allocations among recovery mechanisms do not directly
% track this changing relationship.
Recent work has begun to automate this role. AutoSERL reuses a single
demonstration to guide exploration and recover from predefined failure regions
\cite{liu2026autoserl}, while UniIntervene detects sustained value degradation
and invokes a goal-conditioned recovery policy learned offline
\cite{deng2026uniintervene}. These methods provide useful automated assistance,
but the assistance required by the RL policy changes as its competence
evolves during online RL. Early in learning, intervention can prevent repetitive
failures and provide useful experience; later, excessive assistance can limit
the autonomous execution needed to further improve the RL policy~\cite{mo2026ohp,zhao2025real, xue2026worldsample}. Intervention
strategies based on offline estimates or fixed decision rules can therefore
become mismatched to the current policy.

% We view this challenge as a control-allocation problem. Autonomous execution
% and different intervention behaviors are alternative ways of acting at a given
% stage of training, and their relative values should be learned in a shared
% decision space. Because the RL policy changes throughout online RL, this
% allocation policy must also be updated online. We therefore ask:
% Can an intervention agent learn a unified policy that adaptively
% allocates control between autonomous and assisted behaviors as the RL policy
% improves online?

What is missing is an intervention agent that learns to allocate control
between autonomous execution and heterogeneous assisted behaviors as the task
policy improves online. This requires learning their relative values online
while keeping control allocation responsive to the policy's evolving
competence. We therefore ask: \emph{Can an intervention agent learn a unified
control-allocation policy that adapts as the RL policy improves during
online RL?}

\begin{figure}[t]
    \centering
    \includegraphics[width=\columnwidth]{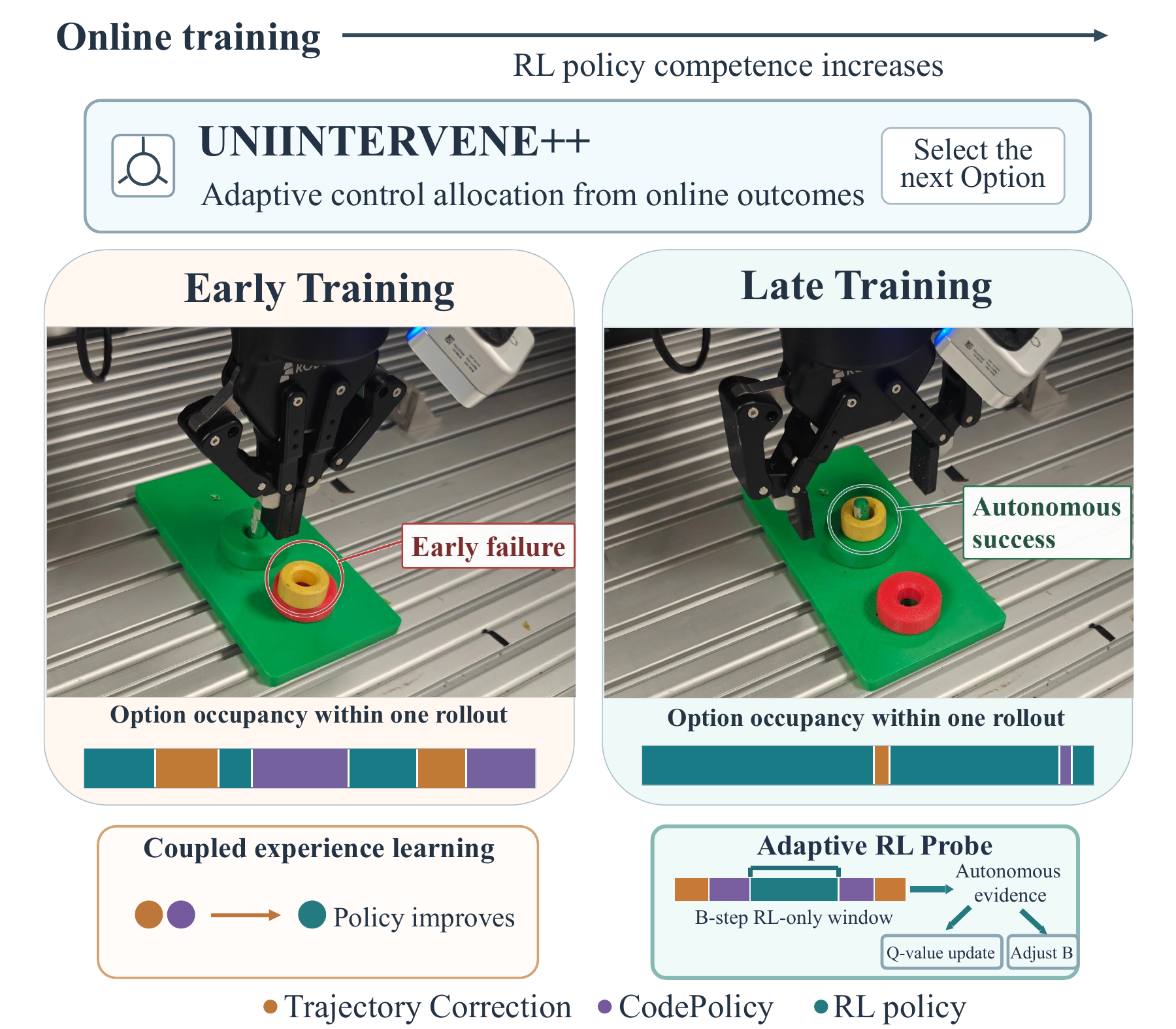}
    \caption{Competence-adaptive control allocation in \method. Illustrative Ring Transfer rollouts show a shift from early failures with substantial use of Trajectory Correction and CodePolicy to later autonomous success with predominantly RL policy control. Colored bars indicate Option occupancy within each rollout (teal: RL policy; orange: Trajectory Correction; purple: CodePolicy). Assisted experience improves the RL policy, while Adaptive RL Probe reserves unassisted windows to assess its evolving competence and guide subsequent allocation.}
    \label{fig:fig1}
    \vspace{-0.4cm}
\end{figure}

To address this gap, we propose \method, an adaptive intervention agent that
learns to allocate control between autonomous execution and heterogeneous
assisted behaviors during online RL. Specifically, to compare autonomous
execution and different forms of assistance within a unified decision process,
\emph{Unified Intervention Policy Learning} places the evolving RL policy,
demonstration-guided trajectory correction, and a task-structured CodePolicy in
a shared decision space and learns their relative values online. To prevent
initially effective assistance from obscuring the RL policy's evolving
competence, \emph{Competence-Adaptive Intervention}, as illustrated in Fig.~\ref{fig:fig1}, periodically probes the RL
policy through unassisted execution, providing direct evidence of its current
competence and enabling control allocation to progressively shift from
assistance-heavy to autonomy-heavy execution. Finally, to ensure that
intervention also contributes to improving the RL policy,
\emph{Coupled Experience Learning} uses assisted experience to train the task
policy, whose evolving outcomes in turn reshape future intervention decisions.
In this way, \method jointly determines when to intervene, how to intervene,
and when to return control as the RL policy improves, without requiring a
separate offline intervention-learning stage. Across five sparse-reward
real-world manipulation tasks, \method achieves an average success rate of
89.67\%, exceeding the best baseline by 6 percentage points, while reducing
the human intervention rate to 0.77\%. Its control allocation progressively
shifts toward the RL policy as autonomous competence improves.

The main contributions are:
\begin{itemize}

\item We formulate online intervention as a unified control-allocation problem,
placing autonomous execution and heterogeneous assisted behaviors in a shared
decision space and learning their relative values online.

\item We develop competence-adaptive intervention with coupled experience
learning, which keeps control allocation responsive to the RL policy's
evolving competence while using assisted experience to improve the policy and
reshape subsequent intervention decisions.

\item Extensive real-world experiments across five manipulation tasks show that
\method achieves 89.67\% average success, outperforming the best baseline by
6 percentage points, while reducing the human intervention rate to 0.77\%.

\end{itemize}

% The main contributions are:
% \begin{itemize}
% \item A unified intervention-policy formulation that places autonomous execution and heterogeneous intervention behaviors in a shared decision space and learns their relative values online.

% \item A competence-adaptive intervention-allocation mechanism that adjusts control according to the evolving capability of the RL policy, naturally shifting from assistance-heavy to autonomy-heavy execution.

% \item Real-robot evidence across five manipulation tasks showing that the learned intervention policy achieves 89.65\% average success with only 0.77\% human intervention.

% \end{itemize}

% TODO: Replace the current Option-scheduler figure with a full-width visual
% thesis showing the RL policy and intervention agent learning online, the
% three decisions (step in / intervene / step out), and zero-additional-cost
% insertion into a new target task.

%%%%%%%%%%%%%%%%%%%%%%%%%%%%%%%%%%%%%%%%%%%%%%%%%%%%%%%%%%%%%%%%%%%%%%%%%%%%%%%%
\begin{figure*}[t]
    \centering
    \includegraphics[width=\textwidth]{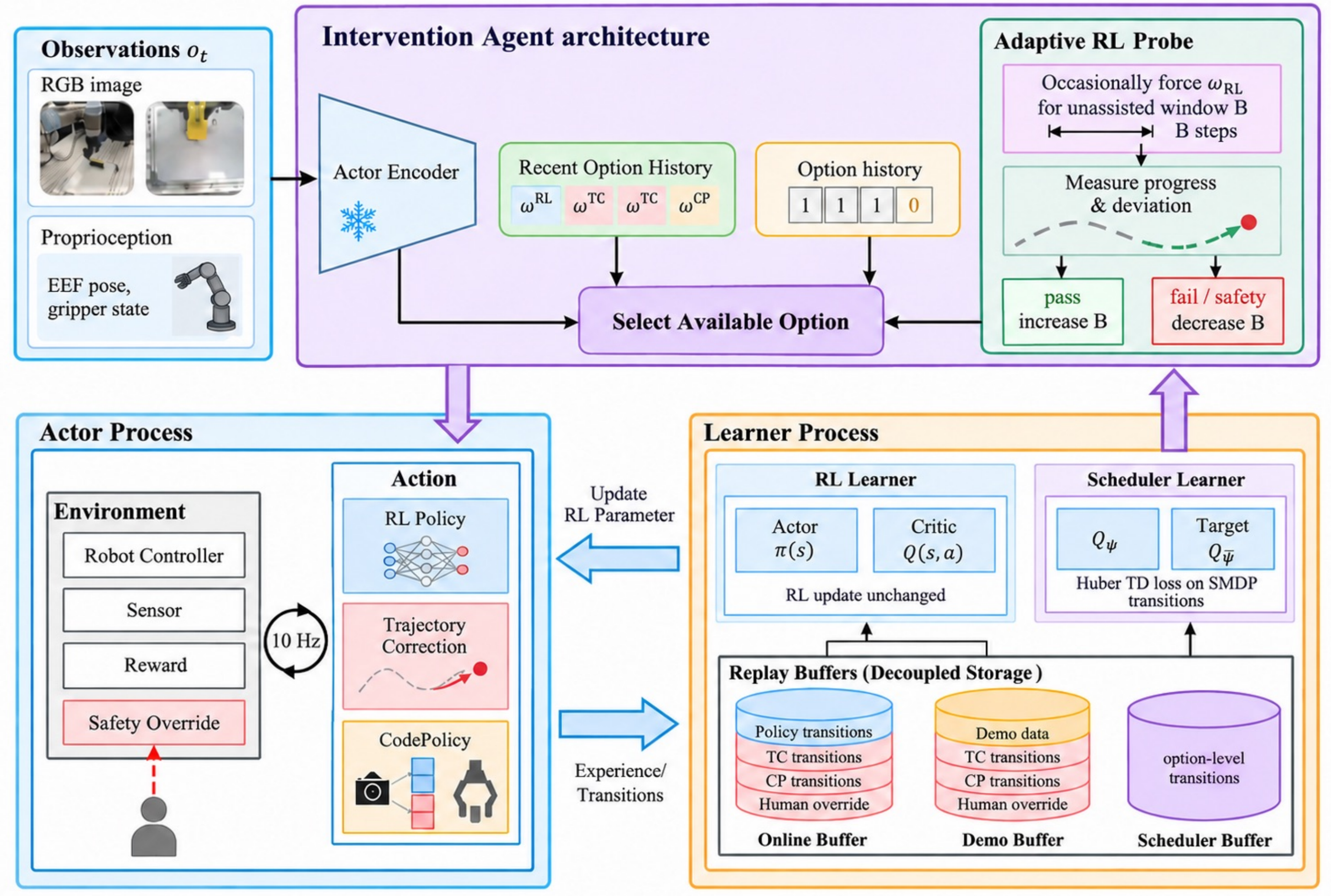}
    \caption{Overview of \method. At each Option boundary, the availability-masked intervention agent selects among the evolving RL policy, trajectory correction, and CodePolicy. Robot-control transitions update the RL policy, while Option-completion transitions update the intervention agent. Adaptive RL Probe preserves direct evidence of the RL policy’s current competence.}
    \label{fig:framework_overview}
    \vspace{-0.4cm}
\end{figure*}
\section{Related Work}

\subsection{Human-in-the-Loop Reinforcement Learning}
Human guidance can make physical policy learning tractable by providing targeted interventions when autonomous exploration becomes unsafe or unproductive.
Interactive imitation learning established this data-collection principle:
DAgger queries expert actions on learner-induced states
\cite{ross2011dagger}; HG-DAgger gives the supervisor control of the takeover
gate, whereas ThriftyDAgger formalizes intervention using novelty and risk
\cite{kelly2019hgdagger,hoque2022thriftydagger}. For robot deployment, Sirius
treats takeovers as both corrective labels and evidence of human trust, then
reweights the collected experience for subsequent policy updates
\cite{liu2025robot}. These supervised approaches mitigate covariate shift but
primarily optimize imitation of corrective behavior.

Reward-driven methods use intervention data beyond behavioral cloning. RLIF
turns the onset of an intervention into a negative reward and applies
off-policy RL, relaxing the assumption that corrective actions are
near-optimal \cite{luo2024rlif}. In parallel, RLPD and SERL provide
sample-efficient off-policy learning that mixes prior data with online
interaction \cite{ball2023rlpd,luo2024serl}. HIL-SERL combines this foundation
with human takeovers, storing corrective transitions in both demonstration and
online replay buffers \cite{luo2025hilserl}. E2HiL instead selects replay
samples by their estimated effect on policy entropy to use costly interventions
more efficiently \cite{deng2026e2hil}. Together, these systems show that
intervention can shape
where data are collected, which corrective actions are executed, and how the
resulting transitions influence learning. They nevertheless generally keep the
operator responsible for monitoring, intervention timing, and corrective
action generation. Our work retains human takeover as an external safety
backstop, but transfers intervention timing, corrective action selection, and
control return to an intervention agent that learns online together with the
RL policy.

\subsection{Automated Intervention and Recovery}

Robust execution is often obtained by pairing a nominal controller with a
localized fallback. Plan patching learns a model-free policy from
demonstrations only where an approximate planner's transition model fails
\cite{lagrassa2020learning}, while manifold-return methods guide a policy from
out-of-distribution observations back toward its demonstration distribution
\cite{reichlin2022back}. In safe RL, Recovery RL and LS3 learn recovery
zones or safe sets from offline experience to constrain subsequent exploration
\cite{thananjeyan2021recovery,wilcox2022ls3}. MetaReSkill instead discovers
failure modes in simulation and uses model-predictive meta-reasoning to allocate
a limited training budget among failure-specific recovery policies
\cite{vats2023efficient}. Contact-mode transition models have also been used
to select among nominal skills and heuristically generated recoveries
\cite{wang2019learning}.

RecoveryChaining learns a local model-free recovery policy in a hybrid action
space containing primitive actions and terminal nominal Options
\cite{vats2025recoverychaining}. AutoSERL automates training-time recovery from
a single successful demonstration \cite{liu2026autoserl}. UniIntervene is the
closest value-based intervention formulation: it learns intervention behavior
offline and uses value degradation to trigger recovery
\cite{deng2026uniintervene}. Its learned values and supporting modules are fixed
before the target online learning process, however, and it does not formulate
control return as a decision that must adapt to the improving RL policy.
\method{} moves the intervention learner into the online loop and learns entry,
corrective behavior, and exit from the same evolving value comparison.

%%%%%%%%%%%%%%%%%%%%%%%%%%%%%%%%%%%%%%%%%%%%%%%%%%%%%%%%%%%%%%%%%%%%%%%%%%%%%%%%

\section{Method}

\subsection{Problem Formulation and Framework Overview}

We formulate the robotic manipulation task as a Markov decision process (MDP) \(\mathcal{M}=(\mathcal{S},\mathcal{A},\rho,P,r,\gamma)\), where \(s_t\in\mathcal{S}\) denotes the state, \(o_t\) the policy observation available at \(s_t\), \(a_t\in\mathcal{A}\) the action, \(r_t\) the task reward, and \(\gamma\) the discount factor. We build on HIL-SERL \cite{luo2025hilserl}, retaining its off-policy RL objective and update rule unchanged. Let \(\pi_\phi(a_t\mid o_t)\) denote the RL policy trained within this framework.
% Our contribution does not alter its RL objective. Instead, we formulate a learning-based intervention agent as an availability-masked SMDP scheduler that dynamically allocates control among the available execution modes. The agent allocates online interaction among the evolving RL policy and two complementary sources of automated assistance. We use the term scheduler hereafter to refer to this learned intervention agent.
Without altering the RL policy objective, we introduce \method, a unified online framework that learns an adaptive intervention agent to dynamically allocate control among the evolving RL policy and two complementary sources of automated assistance, thereby determining when to intervene, which form of assistance to invoke, and when to return control to the policy as it improves. As illustrated in Fig.~\ref{fig:framework_overview}, at each high-level decision boundary, the agent observes the current scene and recent execution outcomes, selects one of the currently available execution modes, and allows the selected controller to interact with the environment until its termination condition is reached. The resulting robot-control transitions are used to continue training the RL policy, while the result of each completed execution mode is used to update the high-level intervention agent.
We represent the available modes of execution as the Option set
\begin{equation}
    \Omega=\{\omega_{\mathrm{RL}},\omega_{\mathrm{TC}},
    \omega_{\mathrm{CP}}\},
    \label{eq:option_set}
\end{equation}
where each $\omega\in\Omega$ is a closed-loop, temporally extended controller
with controller-specific initiation and termination rules
\cite{sutton1999between}: $\omega_{\mathrm{RL}}$ executes the current policy
$\pi_\phi$ for a bounded horizon; $\omega_{\mathrm{TC}}$ feedback-controls the
TCP toward a reachable point on a successful demonstration trajectory; and
$\omega_{\mathrm{CP}}$ composes task-conditioned motion, gripper, and hold
primitives using estimated object poses. The latter two Options are available
only when a valid connection target or feasible plan exists, respectively.
Each Option terminates upon controller-specific completion, budget exhaustion,
episode termination, or an external safety override. Because perception and
primitive execution are fallible, completion of a CodePolicy plan does not
necessarily imply task success.  Selecting $\omega_{\mathrm{RL}}$ is an abstention decision that leaves control with the RL policy, whereas selecting $\omega_{\mathrm{TC}}$ or $\omega_{\mathrm{CP}}$ invokes the corresponding automated intervention.  At high-level decision $n$, an Option begins at robot-control time $t_n$, executes for $k_n$ control steps, and terminates at $t_{n+1}=t_n+k_n$.  The resulting control-allocation problem is therefore a semi-Markov decision process (SMDP)~\cite{janssen2013semi}, rather than a step-wise controller switch.  Human takeover remains an external safety override and is not treated as a fourth Option.

\subsection{Unified Intervention Policy Learning}

We implement the intervention agent as a Double-DQN scheduler with parameters
$\psi$. The agent observes the current scene and recent consequences of its
control choices through the scheduler state
\begin{equation}
    x_n=\left[f_{\bar{\phi}}(o_{t_n}),
    h_{n-H:n-1},v_n\right],
    \label{eq:scheduler_state}
\end{equation}
where $f_{\bar{\phi}}$ is a frozen snapshot of the RL policy's visual encoder,
$h_{n-H:n-1}$ contains the most recent $H$ completed-Option summaries, and
$v_n$ indicates which history entries are valid. Freezing the encoder prevents
the scheduler input from drifting with RL policy updates. Each history entry
records the selected Option, its duration and scheduler reward, end-effector
translation and rotation, path length, and force statistics.

Not every Option is available in every state. We define an availability mask
$m_n\in\{0,1\}^{|\Omega|}$, where $m_n(\omega)=1$ indicates that Option
$\omega$ satisfies its initiation condition. The RL Option is normally
available; trajectory correction requires a valid local reference target, and
CodePolicy requires valid perceptual inputs and a feasible task-conditioned
plan. The mask is used for exploration, greedy selection, and bootstrap target
computation. The greedy selection is
\begin{equation}
    \omega_n=\underset{\omega:\,m_n(\omega)=1}{\arg\max}
    Q_{\psi}(x_n,\omega),
    \label{eq:masked_selection}
\end{equation}
where $Q_{\psi}$ is the scheduler action-value function; an availability-aware $\epsilon$-greedy rule is used during training.

After Option $\omega_n$ executes for $k_n$ robot-control steps, the scheduler
receives
\begin{align}
    R_n^{\mathrm{sched}}
    ={}&\sum_{j=0}^{k_n-1}\gamma^j r_{t_n+j}
    -c_{\mathrm{TC}}\mathbf{1}\{\omega_n=\omega_{\mathrm{TC}}\}
    \nonumber\\
    &-c_{\mathrm{CP}}\mathbf{1}\{\omega_n=\omega_{\mathrm{CP}}\}
    -c_{\mathrm{dur}}\min\!\left(
    \frac{k_n}{K_{\mathrm{norm}}},1\right),
    \label{eq:scheduler_reward}
\end{align}
where $\mathbf{1}\{\cdot\}$ is the indicator function,
$c_{\mathrm{TC}}$ and $c_{\mathrm{CP}}$ penalize calls to the two assisted
Options, $c_{\mathrm{dur}}$ penalizes prolonged execution, and
$K_{\mathrm{norm}}$ normalizes Option duration. These costs discourage
unnecessary or prolonged assistance while preserving task return.

We learn $Q_{\psi}$ using Double DQN \cite{vanhasselt2016deep}. At the
successor state, the online network selects
\begin{equation}
    \omega_n^{*}=
    \underset{\omega:\,m_{n+1}(\omega)=1}{\arg\max}
    Q_{\psi}(x_{n+1},\omega),
    \label{eq:next_option}
\end{equation}
and the target network $Q_{\bar{\psi}}$ evaluates it using the SMDP target
\begin{equation}
    y_n=R_n^{\mathrm{sched}}+b_n\gamma^{k_n}
    Q_{\bar{\psi}}(x_{n+1},\omega_n^{*}),
    \label{eq:smdp_target}
\end{equation}
where $\bar{\psi}$ denotes the target-network parameters and $b_n=0$ for a
terminal transition and $b_n=1$ otherwise. Each completed Option produces an
SMDP transition stored in scheduler replay for the Double-DQN update. The duration-dependent discount $\gamma^{k_n}$
places Options of different durations on a common temporal scale. We optimize
the scheduler with the Huber temporal-difference loss.

% Publication-ready grouped three-line table.
% Requires booktabs, makecell, and multirow.
% The first success fraction in each source cell is used as SR.
% Secondary checkpoint entries are intentionally omitted pending clarification:
% HIL-SERL/Tube: 4/10 = 40%; HIL-SERL/USB: 42/60 = 70%;
% AutoSERL/Board: 17/60 = 28.3% (or 41.7% if "+8" means 8 successes).
% AutoSERL/Tube HIR=0.001 is treated as a fraction and shown as 0.10%.
\begin{table*}[t]
    \caption{Real-world manipulation results across five tasks. Bold and underlined values indicate the best and second-best
results, respectively; ties share the same formatting. Higher SR and lower
HIR are better.}
    \label{tab:task-results}
    \centering
    \begingroup
    \footnotesize
    \setlength{\tabcolsep}{0.8pt}
    \renewcommand{\arraystretch}{1.16}
    \begin{tabular*}{\textwidth}{@{}w{l}{7em}*{12}{w{c}{\dimexpr(\textwidth-7em-24\tabcolsep)/12\relax}}@{}}
        \toprule
        \multirow{3}{*}{\textbf{Method}}
        & \multicolumn{2}{c}{\textbf{Pick and Place}}
        & \multicolumn{6}{c}{\textbf{Contact rich}}
        & \multicolumn{2}{c}{\textbf{Deformable Object}}
        & \multicolumn{2}{c}{\textbf{Average}} \\
        \cmidrule(lr){2-3}\cmidrule(lr){4-9}\cmidrule(lr){10-11}
        & \multicolumn{2}{c}{Ring Transfer}
        & \multicolumn{2}{c}{\makecell[c]{Tube  Insertion}}
        & \multicolumn{2}{c}{\makecell[c]{USB  Insertion}}
        & \multicolumn{2}{c}{\makecell[c]{Wipe  Whiteboard}}
        & \multicolumn{2}{c}{\makecell[c]{Towel  Folding}}
        & \multicolumn{2}{c}{} \\
        \cmidrule(lr){2-3}\cmidrule(lr){4-5}\cmidrule(lr){6-7}
        \cmidrule(lr){8-9}\cmidrule(lr){10-11}
        & SR$\uparrow$ & HIR$\downarrow$
        & SR$\uparrow$ & HIR$\downarrow$
        & SR$\uparrow$ & HIR$\downarrow$
        & SR$\uparrow$ & HIR$\downarrow$
        & SR$\uparrow$ & HIR$\downarrow$
        & SR$\uparrow$ & HIR$\downarrow$ \\
        \midrule
        HIL-SERL
        & 83.33 & 39.15
        & \underline{85.00} & 39.65
        & \underline{90.00} & 51.68
        & 85.00 & 63.11
        & 75.00 & 76.57
        & \underline{83.67} & 54.03 \\
        AutoSERL
        & 53.33 & \underline{17.07}
        & 80.00& \textbf{0.10}
        & 88.33 & \underline{3.91}
        & 80.00 & 16.80
        & 66.67 & 33.55
        & 73.67 & \underline{14.29} \\
        UniIntervene
        & \underline{85.00} & 25.32
        & 73.33 & 15.80
        & \underline{90.00} & 9.90
        & \textbf{90.00} & \underline{10.90}
        & \underline{80.00} & \underline{32.30}
        & \underline{83.67} & 18.84 \\
        \midrule
        \textbf{UniIntervene++}
        & \textbf{88.33} & \textbf{0.30}
        & \textbf{91.67} & \underline{2.58}
        & \textbf{96.67} & \textbf{0.96}
        & \underline{86.67} & \textbf{0.01}
        & \textbf{85.00} & \textbf{0.00}
        & \textbf{89.67} & \textbf{0.77} \\
        \bottomrule
    \end{tabular*}
    \endgroup
\end{table*}

\subsection{Competence-Adaptive Intervention}
% \textbf{Adaptive autonomy probing:}
A scheduler can initially favor a reliable automated controller and thereby
deprive the improving RL policy of both execution opportunities and current
competence estimates.  As an internal behavior-selection mechanism of the
intervention agent, Adaptive RL Probe periodically reserves a contiguous
unassisted window of $B$ robot-control steps.  At each Option boundary within
this window, it temporarily overrides the learned selection rule in
Eq.~\eqref{eq:masked_selection} and forces $\omega_{\mathrm{RL}}$.  The probe
introduces neither an additional scheduler action nor a separate reward or
value function.  Each forced RL Option retains the ordinary Option horizon,
produces a standard SMDP transition with the reward in
Eq.~\eqref{eq:scheduler_reward}, and is stored in scheduler replay.  Thus, one
probe window can produce multiple consecutive RL Option transitions.  Because
the executed Option is $\omega_{\mathrm{RL}}$, its scheduler reward contains no
trajectory-correction or CodePolicy cost.  The probe adapts $B$ from measured
execution evidence rather than equating assisted episode success with the RL
policy's competence.

Let $\Delta\ell_E(B)$ be the demonstration's reference arc-length increment
over the probed segment.  We project the measured TCP poses monotonically onto
the reference path and denote their projected progress by
$\Delta\ell_{\mathrm{RL}}$.  For a segment containing sufficient reference
motion, a completed probe passes when
\begin{equation}
    \Delta\ell_{\mathrm{RL}}\geq\eta\Delta\ell_E(B),
    \qquad d_{\max}\leq d_{\mathrm{allow}},
    \label{eq:probe_pass}
\end{equation}
where $\eta$ is the required relative progress and $d_{\max}$ is the maximum
deviation from the reference path.  Projected arc-length progress is used
instead of unconstrained end-effector travel, so motion in an irrelevant
direction cannot satisfy the probe.  When the reference segment is nearly
stationary, the probe instead evaluates a tighter local-deviation condition
and does not require artificial motion.

Task completion also passes the current probe.  Human takeover, safety
interruption, excessive path deviation, or stalled progress can end it early.
After a configured number of consecutive passes, the unassisted window is
increased; deviation or safety failures shorten it, while other failures clear
the pass streak.  Probe-specific progress, deviation, and pass/fail signals are
used only to update this curriculum and are not added to
$R_n^{\mathrm{sched}}$.  A probe pass is not counted as an independent task success.

% \subsection{Coupled Experience Learning}

% The scheduler and RL learner are coupled through experience, but retain
% separate networks, optimizers, and replay buffers.  All robot-control
% transitions, regardless of whether they were generated by the RL policy, an
% automated Option, or human override, enter the online replay buffer
% $\mathcal{D}_{\mathrm{online}}$.  The demonstration buffer
% $\mathcal{D}_{\mathrm{demo}}$ contains the initial demonstrations and all
% transitions generated by trajectory correction, CodePolicy, and human
% override.  The unchanged HIL-SERL learner
% draws mixed off-policy batches from these two buffers.  Automated assistance
% therefore affects the RL policy through replayed transitions; we introduce
% no additional behavior-cloning or policy-distillation loss.  Eligibility for
% the demonstration buffer is a configured source rule, rather than a learned
% quality or success filter.

% The RL policy and scheduler are not trained end-to-end by shared gradients.
% Their interaction instead forms a coupled data loop operating at robot-control
% and Option timescales:
% automated Options assist difficult portions and generate usable experience;
% off-policy replay improves the RL policy; and new unassisted rollouts
% subsequently change the scheduler's empirical value estimates.  Adaptive
% probing preserves the unassisted policy evidence needed to keep this loop
% responsive as the RL policy evolves.

\subsection{Coupled Experience Learning}

The scheduler and RL learner are coupled through experience while retaining
separate networks, optimizers, and replay buffers. Let
$\xi_t=(o_t,a_t,r_t,o_{t+1})$ denote a robot-control transition and $z_t$ its
execution source. All transitions enter the online replay buffer, while
transitions generated by automated assistance or human override are
additionally routed to the demonstration buffer:
\begin{equation}
\begin{gathered}
\mathcal{D}_{\mathrm{online}}
\leftarrow
\mathcal{D}_{\mathrm{online}}\cup\{\xi_t\},\\
\mathcal{D}_{\mathrm{demo}}
\leftarrow
\mathcal{D}_{\mathrm{demo}}\cup\{\xi_t\},
\quad
z_t\in\{\mathrm{TC},\mathrm{CP},\mathrm{Human}\}.
\end{gathered}
\label{eq:experience_routing}
\end{equation}
Here, $\mathcal{D}_{\mathrm{demo}}$ is initialized with the task
demonstrations, and $\mathrm{TC}$ and $\mathrm{CP}$ denote Trajectory
Correction and CodePolicy, respectively. The unchanged HIL-SERL learner draws
mixed off-policy batches from $\mathcal{D}_{\mathrm{online}}$ and
$\mathcal{D}_{\mathrm{demo}}$. Thus, assisted experience directly contributes to RL policy learning without introducing an additional behavior-cloning or policy-distillation objective.

The RL policy and scheduler are not coupled through shared gradients.
Instead, they form a coupled data loop across robot-control and Option
timescales: automated Options generate useful experience, off-policy replay
improves the RL policy, and the resulting policy produces new execution
outcomes that update the scheduler's value estimates. Adaptive RL Probe
preserves direct evidence of autonomous competence, keeping this loop
responsive as the RL policy evolves.

%%%%%%%%%%%%%%%%%%%%%%%%%%%%%%%%%%%%%%%%%%%%%%%%%%%%%%%%%%%%%%%%%%%%%%%%%%%%%%%%
\begin{figure}[t]
    \centering
    \includegraphics[
        width=\linewidth,
        height=0.75\textheight,
        keepaspectratio
    ]{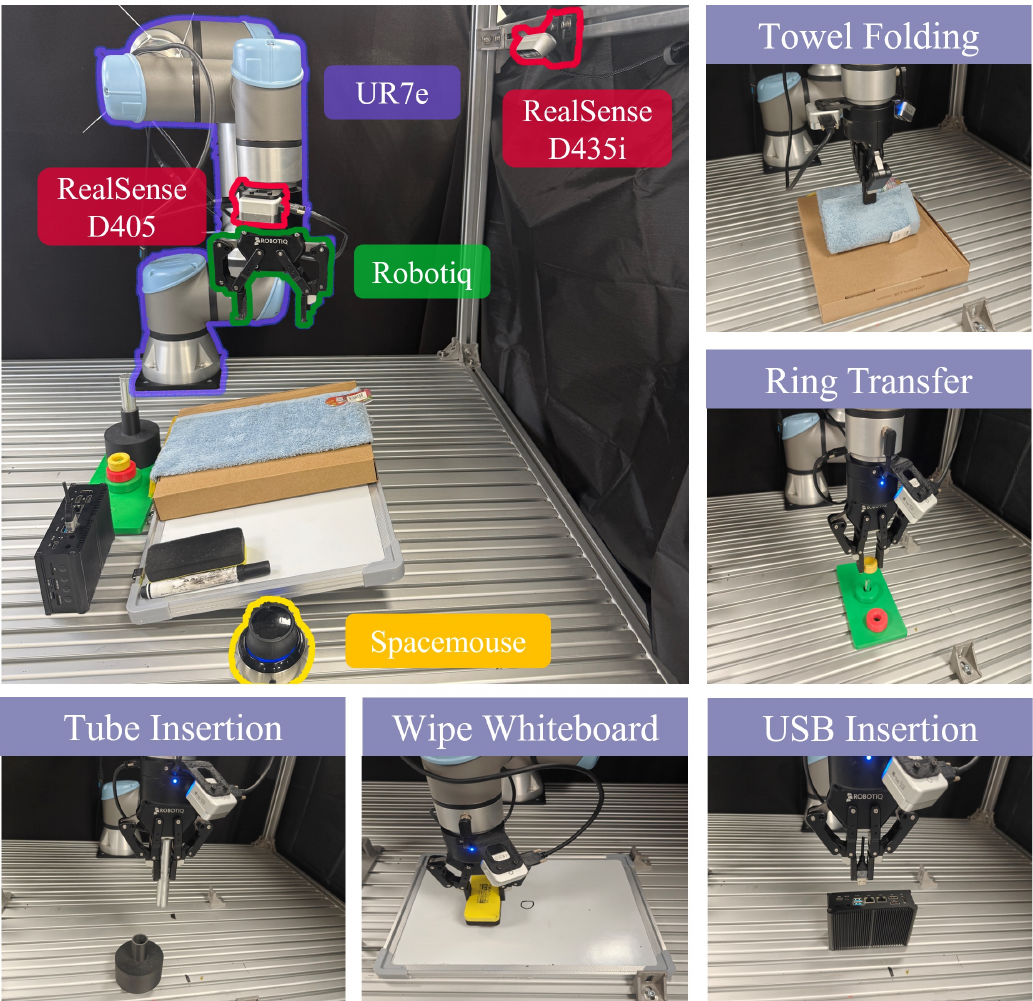}
    \caption{UR7e setup for five manipulation tasks.}
    \label{fig:experimental-setup}
    \vspace{-0.4cm}
\end{figure}
\section{Experiments}

We conduct real-world robotic manipulation experiments to answer three key
questions:
(1) Can \method{} improve task success while reducing human interventions
compared with HiL-RL methods (Sec.~\ref{subsec:Experiments result and Analysis})?
(2) Does \method{} adapt its intervention timing, Option selection, and control
handoff as the RL policy evolves, enabling progressive autonomy (Sec.~\ref{subsec:Experiments result and Analysis})?
(3) Does \method{} outperform fixed-prior stochastic selection, and how does
each source of automated assistance contribute to the final RL policy (Sec.~\ref{subsec:Ablations and Diagnostics})?

\subsection{Experiment Setup}
\label{sec:exp_setup}
\textbf{Real-world benchmark.}
As shown in Fig.~\ref{fig:experimental-setup}, all experiments are conducted on a single UR7e robotic arm equipped with a Robotiq gripper. Human corrective actions and external safety overrides are issued through a 3Dconnexion SpaceMouse.
We evaluate \method{} on Ring Transfer, Tube Insertion, USB Insertion, Wipe Whiteboard, and Towel Folding. The benchmark spans
precision manipulation, contact-rich insertion and interaction, and
deformable-object manipulation. Collectively, these tasks test constrained
extraction and placement, precise alignment, contact-aware
correction, sustained tool--surface interaction, and manipulation under
deformable-object dynamics.

\textbf{Baselines.}
We compare \method{} with three baselines for human-in-the-loop manipulation.
\textbf{HIL-SERL}
\cite{luo2025hilserl} uses an operator to provide corrective actions
during online SAC/RLPD training. \textbf{AutoSERL} \cite{liu2026autoserl}
derives trajectory guidance and automated recovery from a single
demonstration. \textbf{UniIntervene} \cite{deng2026uniintervene} is the offline,
value-based counterpart: it detects sustained value degradation and invokes a
memory-guided, goal-conditioned recovery policy using modules learned before
online deployment. All methods use the same perception inputs, task rewards,
action limits, prior data, and robot-interaction budget wherever
applicable. \method{} retains the same low-level learner, but learns the
intervention agent jointly with it online. Human override remains available
as an external safety layer.

\textbf{Evaluation metrics.}
We report success rate (SR) and human
intervention rate (HIR). SR is measured over three evaluation runs
of 20 episodes per task after training. HIR counts human-controlled steps. In addition, we evaluate the RL policy at multiple checkpoints throughout
training and report the duration-weighted control share of each Option,
including the RL policy, CodePolicy, and Trajectory Correction. These
diagnostics assess whether changes in control allocation track the evolving
performance of the RL policy and whether control progressively shifts from
assisted Options to autonomous execution.

\subsection{Experiments Result and Analysis}
\label{subsec:Experiments result and Analysis}
\textbf{Task performance.}
As shown in Table~\ref{tab:task-results}, \method{} achieves the highest average
success rate (SR), reaching $89.67\%$ across the five real-world tasks under
matched training budgets. This is $6$ percentage points higher than the best
baseline average of $83.67\%$, achieved by HIL-SERL and UniIntervene. At the
task level, \method{} outperforms all baselines on Ring Transfer, Tube Insertion, USB Insertion, and Towel Folding by $3.33$, $6.67$,
$6.67$, and $5.00$ percentage points respectively. On Wipe Whiteboard, \method{} achieves $86.67\%$, which is $3.33$ percentage points below UniIntervene while remaining higher than HIL-SERL and AutoSERL. These results
demonstrate that the proposed adaptive intervention agent improves policy learning across precision, contact-rich, and deformable-object manipulation tasks.

Across the five tasks, \method{} reaches the reported performance using
$18{,}000$--$82{,}000$ robot-interaction steps, corresponding to
$29$--$131$ minutes of effective training time. USB Insertion requires the
shortest run, whereas Ring Transfer requires the longest. This places the
reported task performance within practical real-robot interaction budgets.

\begin{figure}[t]
    \centering
    \includegraphics[width=\columnwidth]{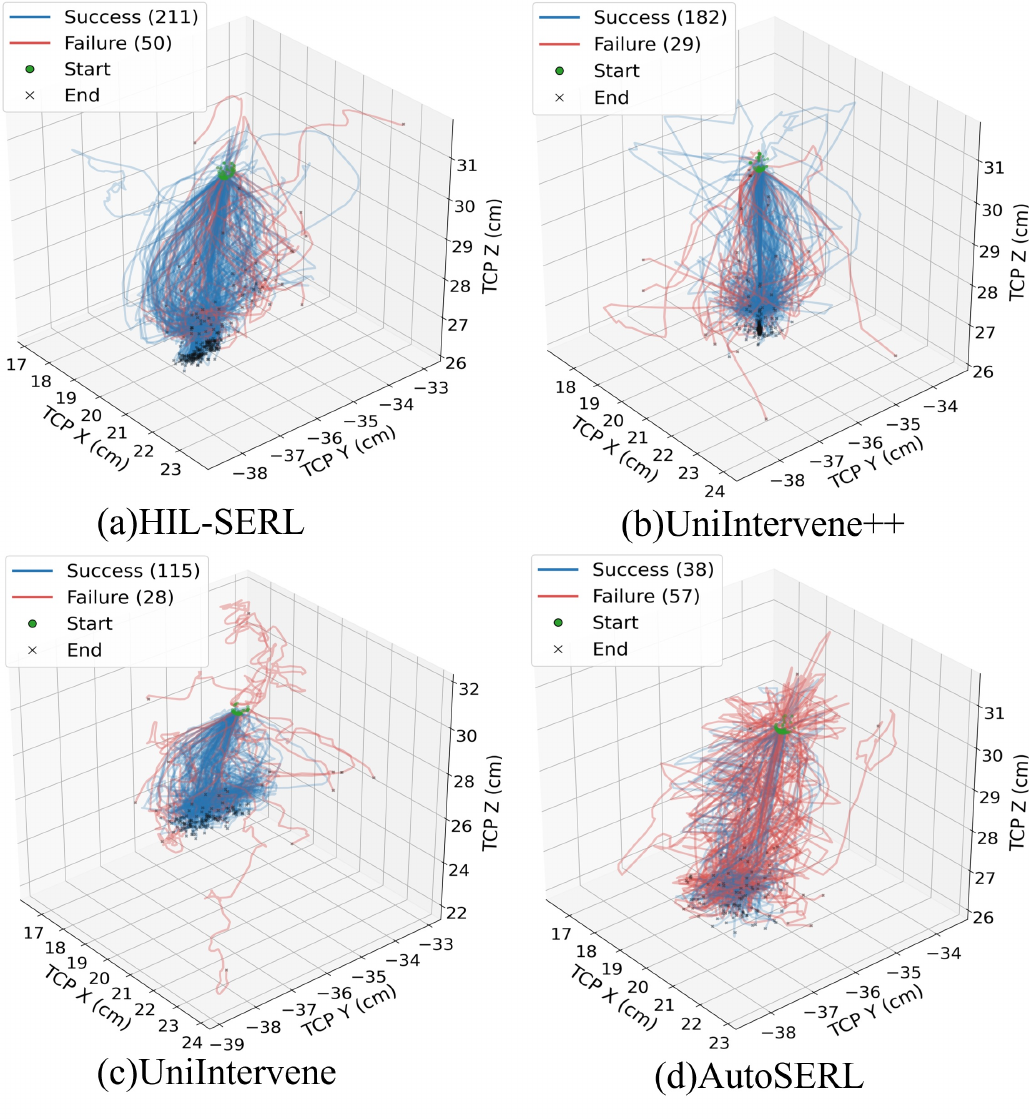}
    \caption{TCP trajectories stored in the replay buffers during the USB Insertion task. Blue and red curves denote successful and failed trajectories, respectively; green circles and black crosses indicate the start and end points. Numbers in parentheses report the number of trajectories in each outcome category. }
    \label{fig:usb-buffer-trajectories}
    \vspace{-0.5cm}
\end{figure}

Using USB Insertion as a representative example,
Fig.~\ref{fig:usb-buffer-trajectories} visualizes the replay-buffer trajectories collected by the four methods. Rather than suppressing autonomous exploration, UniIntervene++
interleaves RL execution with structured corrective behaviors. The
resulting replay buffer therefore combines exploratory experience from
the evolving RL policy with spatially consistent successful trajectories
generated through automated assistance. This balance may provide more
informative training data for off-policy learning while preserving the
policy's opportunity to explore.

\begin{figure}[t]
    \centering
    \includegraphics[width=\columnwidth]{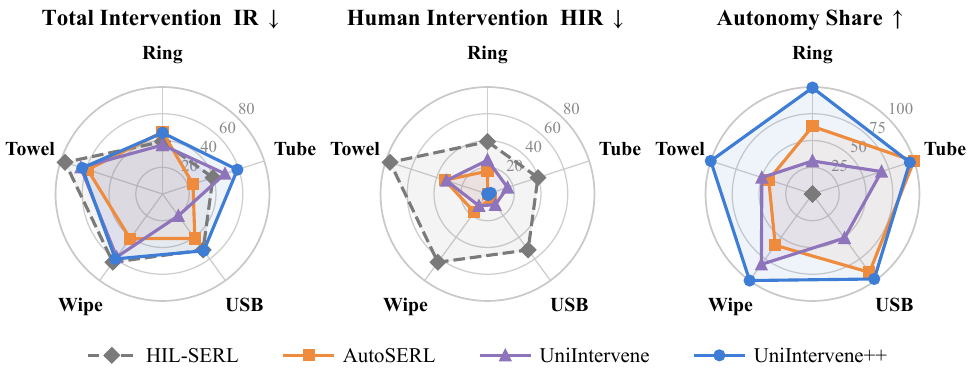}
    \caption{Task-wise intervention decomposition across five real-world
manipulation tasks. IR denotes the fraction of training steps executed under automated or human intervention, while HIR denotes the subset executed under human control. The automated-intervention share is $(\mathrm{IR}-\mathrm{HIR})/\mathrm{IR}$. Arrows indicate the preferred
direction. }
    \label{fig:rad}
    \vspace{-0.6cm}
\end{figure}

\textbf{Human and overall intervention.}
As shown in Fig.~\ref{fig:rad}, \method{} achieves the lowest average
human intervention rate (HIR) in the five tasks, only at $0.77\%$,
compared to $54.03\%$ for HIL-SERL, $14.29\%$ for AutoSERL, and
$18.84\%$ for UniIntervene. Relative to AutoSERL, which has the lowest
HIR among the baselines, \method{} reduces human-controlled steps by
$13.52$ percentage points or $94.6\%$ relative. Although its average
overall intervention rate (IR) is $56.04\%$, $95.62$--$100\%$ of its
intervention steps are executed by automated Options. Thus, the overall
IR reflects the active use of automated assistance rather than the reliance on human takeover.

The differences in HIR reflect how the methods handle failures during
online training. HIL-SERL relies on an operator for routine correction,
resulting in the highest human-control requirement. AutoSERL is designed
to automate intervention using a single successful demonstration
\cite{liu2026autoserl}. However, in our benchmark, its trajectory-based
guidance does not resolve all task-specific failures. On some tasks,
automated intervention alone fails to generate additional
positive-reward online trajectories, making human correction necessary
during training. UniIntervene reduces reliance on human control through
a learned recovery policy, but its intervention modules are learned
offline and remain fixed during target-task online training. Its
remaining human intervention suggests that the fixed recovery behavior
does not cover all failure states encountered by the evolving task
policy. For \method{}, human takeovers occur primarily during early RL
exploration, when the immature policy may enter unsafe states requiring
an external safety override. Routine correction and recovery are instead
handled by Trajectory Correction and CodePolicy. Thus, \method{} reduces human effort by assigning routine
interventions to automated assistance, rather than by minimizing the total amount of assisted execution.

\begin{figure}[t]
    \centering
    \includegraphics[width=\columnwidth]{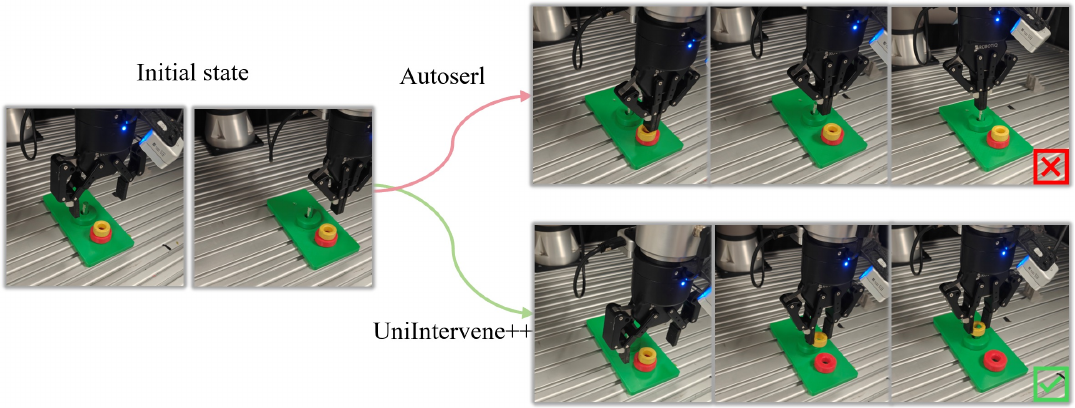}
    \caption{Visualization and comparison on the Ring Transfer task. From the same
initial state, AutoSERL fails to complete the transfer, whereas \method completes the task through adaptive control allocation between autonomous execution and assisted behaviors. This illustrates how competence-adaptive intervention improves robust execution.}
    \label{fig:fig4}
    \vspace{-0.3cm}
\end{figure}

\textbf{Qualitative comparison of recovery behavior.}
Fig.~\ref{fig:fig4} illustrates a failure mode that AutoSERL does not resolve. After a misaligned approach, the gripper closes without grasping the ring. In this rollout, AutoSERL returns the end effector to the demonstrated path but does not reopen the gripper or retry the grasp. The robot then follows the nominal transport motion without the object; without human takeover, the episode fails. By contrast, \method{} uses task-structured assistance with explicit gripper actions to reopen, realign, and regrasp the ring before continuing the transfer. This qualitative comparison highlights a specific recovery advantage of \method{} over AutoSERL in Ring Transfer: after an empty grasp, its task-structured assistance reopens the gripper and retries the grasp, enabling the transfer to succeed without human intervention.

\textbf{Progressive autonomy.}

\begin{figure}[t]
    \centering
    \includegraphics[width=\columnwidth]{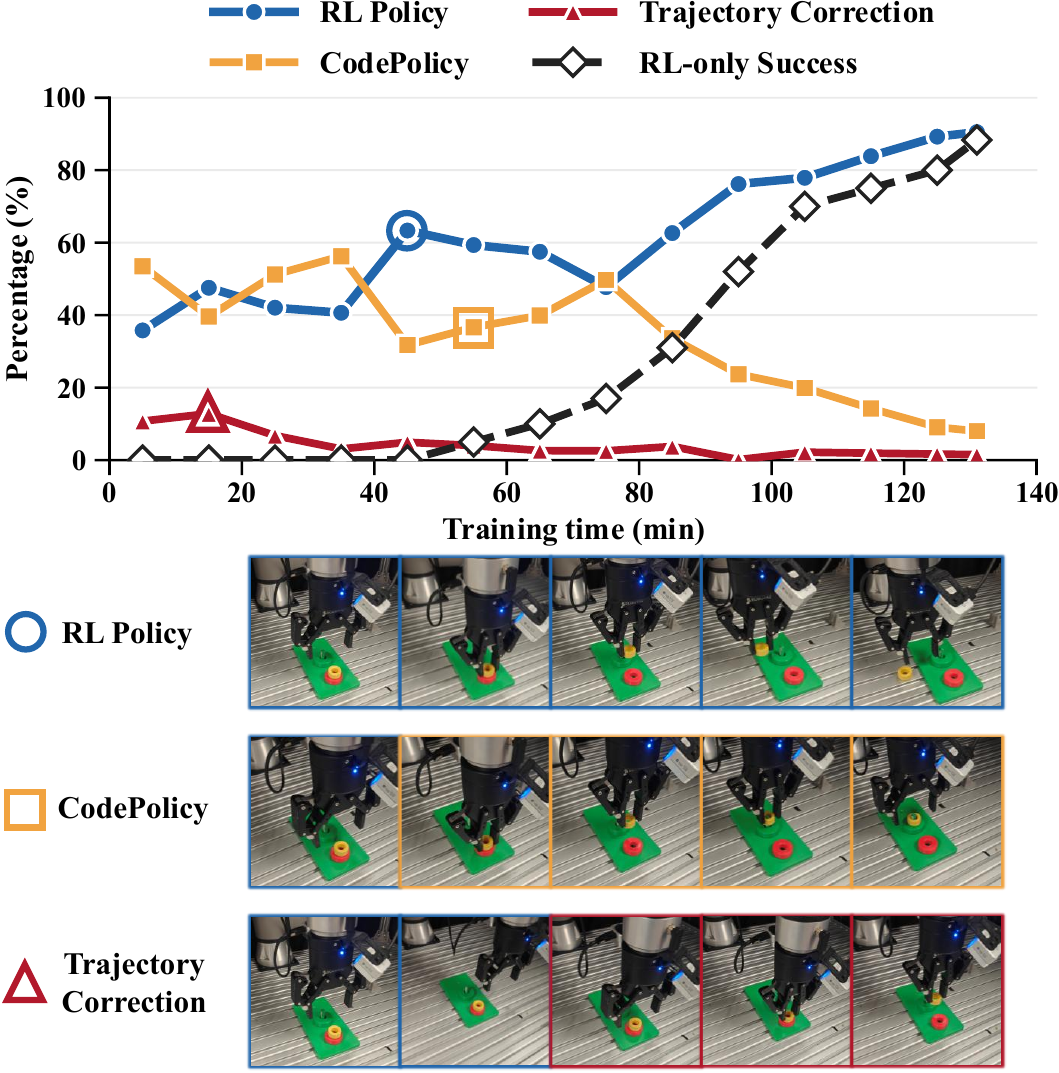}
\caption{Duration-weighted control allocation among the three Options and
RL policy success evaluated at multiple checkpoints during Ring Transfer
training. Within each time window, the control share of an Option is computed
as its accumulated execution duration divided by the total execution duration
of all three Options. The lower rows show representative execution sequences
of the RL policy (circle), CodePolicy (square), and Trajectory Correction
(triangle).}
    \label{fig:scheduler-allocation}
    \vspace{-0.5cm}
\end{figure}

Fig.~\ref{fig:scheduler-allocation} shows that, as RL policy success on Ring
Transfer increases from zero to $88.33\%$, the scheduler raises the RL policy's
control share from $35.78\%$ to $90.51\%$. Meanwhile, the combined allocation
to CodePolicy and Trajectory Correction decreases from $64.22\%$ to $9.49\%$.
This shift demonstrates that the scheduler progressively returns control to the
RL policy as its autonomous competence improves.

The lower rows qualitatively illustrate the three control modes: the RL Option executes the current learned policy, CodePolicy follows a perception-grounded task plan, and Trajectory Correction steers toward a reachable point on a successful demonstration trajectory. These sequences illustrate controller behavior rather than a matched-condition comparison of their performance.

\subsection{Ablations and Diagnostics}
\label{subsec:Ablations and Diagnostics}
We evaluate five configurations on Ring Transfer and Towel Folding to examine
the contribution of each automated assistance Option and the benefit of learned
Option allocation. \textbf{SERL} disables both the intervention agent and human
correction, serving as the fully unassisted baseline.
\textbf{w/o CodePolicy} removes $\omega_{\mathrm{CP}}$, while
\textbf{w/o Trajectory Correction} removes $\omega_{\mathrm{TC}}$ but retains
the trajectory reference required by Adaptive RL Probe.
\textbf{Fixed-prior stochastic scheduler} replaces learned Q-based selection
with fixed sampling probabilities of $0.3$, $0.5$, and $0.2$ for RL,
Trajectory Correction, and CodePolicy, respectively, while retaining the same
Options, feasibility masks, and Adaptive RL Probe.
\textbf{\method{}} uses all three Options and learns their allocation with the
availability-masked SMDP Double-DQN scheduler.
Table~\ref{tab:ablation-results} reports final success rates under the same
training time.
% Single-column ablation table for the two real-robot tasks.
% Requires booktabs and makecell (loaded in root.tex).
\begin{table}[t]
    \caption{Ablation Study on Ring Transfer and Towel Folding}
    \label{tab:ablation-results}
    \centering
    \begingroup
    \footnotesize
    \setlength{\tabcolsep}{3pt}
    \renewcommand{\arraystretch}{1.15}
    \begin{tabular*}{\columnwidth}{@{\extracolsep{\fill}}lcc@{}}
        \toprule
        \textbf{Training configuration}
        & \makecell[c]{\textbf{Ring} \textbf{Transfer}}
        & \makecell[c]{\textbf{Towel} \textbf{Folding}} \\
        & SR(\%) & SR(\%) \\
        \midrule
        \makecell[l]{SERL}
        & 0
        & 0 \\
         w/o CodePolicy
        & 0
        & 0 \\
         w/o Trajectory Correction
        & 50
        & 47 \\

        \makecell[l]{Fixed-prior stochastic scheduler}
        & 75
        & 80 \\
        \midrule
        \textbf{UniIntervene++}
        & \textbf{88.33}
        & \textbf{85} \\
        \bottomrule
    \end{tabular*}

    \vspace{2pt}
    % \begin{minipage}{\columnwidth}
    %     \footnotesize
    %     \raggedright
    %     \textit{Note:} Each entry reports the number of successful trials over
    %     the total number of evaluation trials, followed by the corresponding
    %     success rate. All evaluations use the deterministic task policy with
    %     automatic assistance disabled.
    % \end{minipage}
    \endgroup
    \vspace{-0.5cm}
\end{table}

The full method achieves $88.33\%$ SR on Ring Transfer and $85\%$ on Towel
Folding. Replacing learned Option selection with the fixed-prior scheduler
reduces SR to $75\%$ and $80\%$, respectively, supporting the benefit of
adaptive online allocation when the controller set, feasibility masks, and
probing mechanism are held fixed.
Without CodePolicy, final SR falls to zero on both tasks, matching the unassisted SERL baseline, as these ablated policies are largely unable to acquire positive samples without relying on human intervention. In comparison, removing Trajectory Correction reduces SR to $50\%$ and $47\%$. Under the evaluated budgets, CodePolicy therefore has the
larger measured contribution, with Trajectory Correction providing
complementary assistance. These are system-level ablations because removing an
Option changes both executed assistance and the experience routed to the task
policy. All scheduler-enabled variants retain Adaptive RL Probe, whose
independent contribution is not isolated here.
\section{Conclusion}

We presented \method{}, an adaptive intervention agent that learns when to
intervene, which assistance to use, and when to return control as the task
policy improves online. \method{} places the evolving RL policy, trajectory
correction, and a task-structured CodePolicy in a shared decision space and
learns their relative values online. Competence-adaptive intervention preserves
direct evidence of the RL policy's evolving competence, while coupled
experience learning allows assisted experience to improve the RL policy and
reshape subsequent intervention decisions. Across five real-world manipulation
tasks, \method{} achieved 89.67\% average success, exceeding the best baseline
by 6 percentage points, with only 0.77\% human intervention. On Ring
Transfer, control allocated to the RL policy increased from 35.78\% to 90.51\%
as its independently evaluated success rose from zero to 88.33\%; ablations on
two tasks further support learned allocation over a fixed prior. These results
show that adaptive intervention can reduce operator effort and progressively
return control to the RL policy without a separate offline
intervention-learning stage.

%%%%%%%%%%%%%%%%%%%%%%%%%%%%%%%%%%%%%%%%%%%%%%%%%%%%%%%%%%%%%%%%%%%%%%%%%%%%%%%%
 \section*{Acknowledgment}
Generative AI tools were used for language editing and to generate selected illustrative icons in Fig.~2. These icons are used solely for visualization and do not represent experimental data or results. All technical content, methodology, experiments, analysis, and conclusions were developed and verified by the authors.

\bibliographystyle{IEEEtran}
\bibliography{references}

\end{document}